\documentclass{article}

\usepackage[main, final]{neurips_2026}

\usepackage[utf8]{inputenc} 
\usepackage[T1]{fontenc}    
\usepackage{hyperref}       
\usepackage{url}            
\usepackage{booktabs}       
\usepackage{amsfonts}       
\usepackage{nicefrac}       
\usepackage{microtype}      
\usepackage{xcolor}         

\usepackage{enumitem}
\usepackage{amsmath}
\usepackage{multirow}
\usepackage{graphicx}
\usepackage[misc]{ifsym}

\newcommand{\authorskip}{\hspace{3mm}}
\newcommand{\institutionskip}{\hspace{2mm}}

\title{PointForward: Feedforward Driving Reconstruction through Point-Aligned Representations}

\author{
  Cheng Chi\textsuperscript{1, 4 *} \authorskip
  Xianqi Wang\textsuperscript{2, 1 *} \authorskip
  Hongcheng Luo\textsuperscript{1, 4 * $\dagger$} \authorskip
  Mingfei Tu\textsuperscript{1, 4} \vspace{0.5mm} \\
  \textbf{Gangwei Xu\textsuperscript{5} \authorskip
  Zehan Zhang\textsuperscript{1, 4} \authorskip
  Bing Wang\textsuperscript{1, 4} \authorskip
  Guang Chen\textsuperscript{1, 4}} \vspace{0.5mm} \\
  \textbf{Hangjun Ye\textsuperscript{1, 4} \authorskip
  Sida Peng\textsuperscript{3} \authorskip
  Xin Yang\textsuperscript{5, 6 \Letter} \authorskip
  Haiyang Sun\textsuperscript{1, 4}}
  \vspace{1.5mm} \\
  \small{
  \textsuperscript{1}Xiaomi EV \institutionskip
  \textsuperscript{2}School of AIA, Huazhong University of Science and Technology \institutionskip
  \textsuperscript{3}Zhejiang University} \\
  \small{
  \textsuperscript{4}Beijing Key Laboratory of Intelligent Driving Foundation Models and Simulation} \\
  \small{
  \textsuperscript{5}Huazhong University of Science and Technology \institutionskip
  \textsuperscript{6}Optics Valley Laboratory} \\
  \vspace{1mm}
  \small{\url{https://wm-research.github.io/PointForward}}
}

\begin{document}

\maketitle

\begin{figure}[h]
  \centering
  \includegraphics[width=1.0\textwidth]{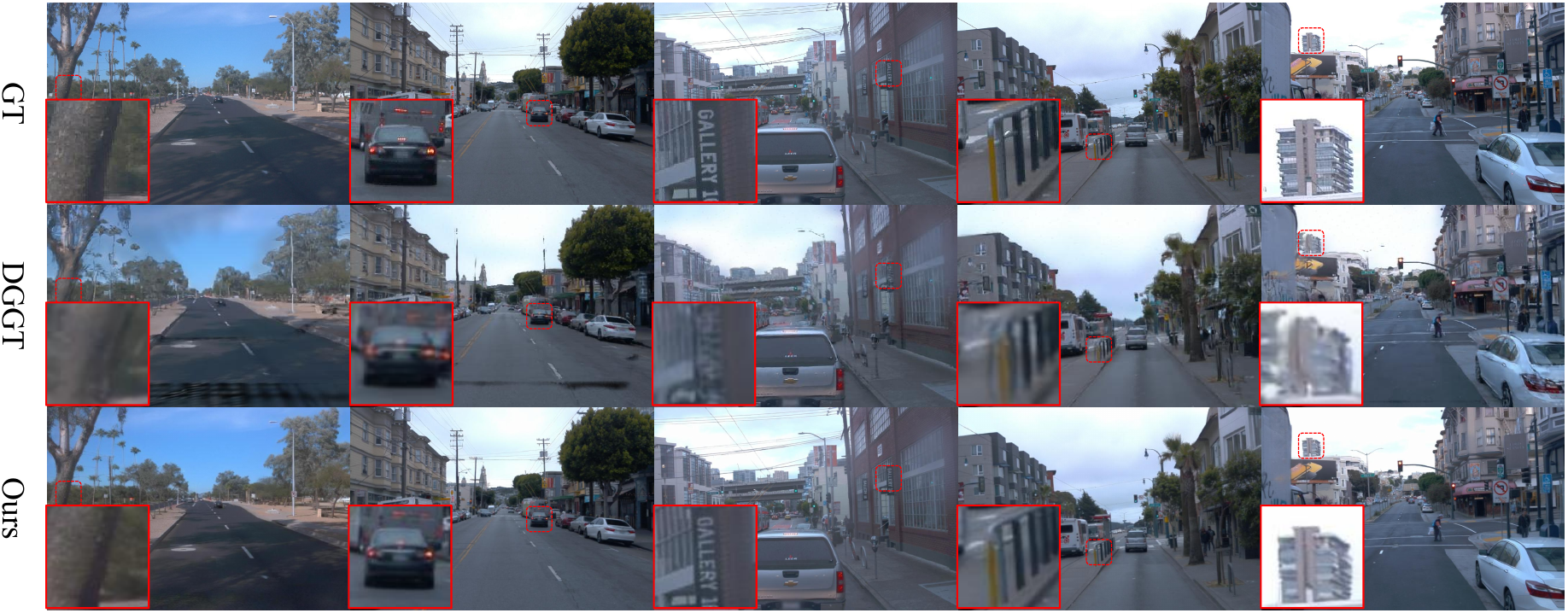}
  \caption{We present \textbf{PointForward}, a feedforward driving reconstruction framework through point-aligned representations. Compared to directly rendering results from the SOTA pixel-aligned method~\cite{chen2025dggt}, it reconstructs fine structures without layering artifacts.}
  \label{fig:teaser}
\end{figure}

\let\thefootnote\relax
\footnotetext[0]{
\small
\textsuperscript{*} Equal contribution, 
\textsuperscript{$\dagger$} Project leader,
\textsuperscript{\Letter} Corresponding author.
}

\begin{abstract}

High-fidelity reconstruction of driving scenes is crucial for autonomous driving. While recent feedforward 3D Gaussian Splatting (3DGS) methods enable fast reconstruction, their per-pixel Gaussian prediction paradigm often suffers from multi-view inconsistency and layering artifacts. Moreover, existing methods often model dynamic instances via dense flow prediction, which lacks explicit cross-view correspondence and instance-level consistency. In this paper, we propose \textbf{PointForward}, a feedforward driving reconstruction framework through point-aligned representations. Unlike pixel-aligned methods, we initialize sparse 3D queries in world space and aggregate multi-view image information via spatial-temporal fusion onto these queries, enforcing explicit cross-view consistency in a single feedforward pass. To handle scene dynamics, we introduce scene graphs that explicitly organize moving instances during reconstruction. By leveraging 3D bounding boxes, our method enables instance-level motion propagation and temporally consistent dynamic representations. Extensive experiments show that PointForward achieves state-of-the-art performance on large-scale driving benchmarks.

\end{abstract}

\section{Introduction}

Accurate reconstruction of driving scenes is fundamental to autonomous driving. Early methods based on Neural Radiance Fields (NeRFs)~\cite{mildenhall2021nerf, pumarola2021d} achieved impressive photorealism, but required hours of per-scene optimization and incurred prohibitive rendering costs. The advent of 3D Gaussian Splatting (3DGS)~\cite{kerbl20233d} dramatically improved efficiency by replacing implicit fields with explicit point-based primitives and a fast differentiable rasterizer. However, early 3DGS-based methods~\cite{chen2024omnire, yan2024street, yang2024deformable} still relied on per-scene optimization, which remained computationally expensive and ill-suited for large-scale real-time deployment.

Recently, feedforward 3DGS methods~\cite{charatan2024pixelsplat, chen2025dggt, chen2024mvsplat, miao2025evolsplat, tan2026ufo, xu2025depthsplat, yang2024storm} have gained traction. By training on large-scale data, these models can infer Gaussians in a single forward pass, reducing reconstruction time from minutes to sub-seconds. Despite this speedup, two structural limitations hinder further improvements in accuracy. First, current methods primarily rely on per-pixel Gaussian prediction: Gaussians are independently predicted from image pixels in each view, yet the same physical point may be projected to slightly different 3D locations across views due to prediction inconsistencies. This leads to redundant Gaussians and multi-view inconsistencies, often manifesting as layering or ghosting artifacts in fine-grained regions. Second, previous feedforward methods model scene dynamics via dense flow prediction~\cite{chen2025dggt, yang2024storm}. They typically rely on 2D dynamic masks to label scene dynamics and then predict per-Gaussian 3D motion flows. 2D dynamic masks cannot fully capture the extent of 3D instances, and dense flow prediction models motion independently at the pixel level, lacking coherent instance-level representations. These limitations motivate us to develop a point-aligned feedforward reconstruction framework.

In this paper, we propose \textbf{PointForward}, a feedforward driving reconstruction framework that goes beyond per-pixel Gaussian prediction by operating on sparse spatial-temporal queries. To build a global point-aligned representation, PointForward first initializes a set of sparse 3D queries directly in world space. These queries are then projected onto all input views to aggregate corresponding image features and geometric cues. Through spatial-temporal fusion, each query is updated into a globally consistent and view-coherent 3D representation. As illustrated in Figure ~\ref{fig:teaser}, PointForward faithfully reconstructs fine structures without duplicate Gaussians, whereas prior pixel-aligned methods often produce noticeable redundancy and artifacts. 

For dynamic modeling, PointForward likewise adopts a point-aligned formulation. Inspired by scene graphs commonly used in per-scene reconstruction~\cite{chen2024omnire, yan2024street}, we explicitly organize dynamic instances and static backgrounds in 3D space using 3D bounding boxes. Queries associated with dynamic instances are transformed into canonical spaces, in which instance motion can be modeled consistently over time. During rendering, these dynamic queries are mapped back to the target world space according to the desired timestamp. This design enables instance-level motion propagation and temporally coherent dynamic representations, avoiding the fragmented motions often produced by dense flow prediction.

We evaluate PointForward on the Waymo Open Dataset~\cite{sun2020scalability} and nuScenes~\cite{caesar2020nuscenes}. Experimental results show that PointForward consistently outperforms prior feedforward methods. On Waymo, PointForward achieves a global PSNR of 28.48 and a dynamic PSNR of 25.01. When directly generalized to nuScenes, it obtains a PSNR of 26.54, with further gains after fine-tuning. These results demonstrate the effectiveness of our point-aligned formulation and scene-graph-based dynamic modeling. Qualitative and Quantitative results further show that PointForward recovers fine details under extrapolated viewpoints while avoiding ghosting artifacts.

Our contributions can be summarized as follows:

\begin{itemize}[leftmargin=*, itemsep=0.5em]
    \item We propose \textbf{PointForward}, a feedforward driving reconstruction framework through point-aligned representations. It aggregates multi-view image information and enforces cross-view consistency in a single forward pass.
    
    \item We introduce a point-aligned representation that replaces per-pixel Gaussian prediction with sparse 3D queries. We further incorporate scene graphs for dynamic modeling, enabling instance-level motion propagation and temporally consistent dynamic reconstruction.
    
    \item We achieve state-of-the-art performance on the Waymo and nuScenes benchmarks, with consistent improvements in both image quality and metric performance.
\end{itemize}

\section{Related Work}

\paragraph{Per-Scene Reconstruction.} 3D Gaussian Splatting (3DGS)~\cite{kerbl20233d} explicitly represents scenes using anisotropic Gaussian primitives and enables real-time high-quality novel view synthesis through differentiable rasterization, laying the foundation for subsequent driving-scene reconstruction methods. However, the original 3DGS is designed for static scenes and cannot directly handle dynamic instances or unbounded street environments. To model scene dynamics, early works introduced implicit temporal representations without explicit supervision. PVG~\cite{chen2026periodic} embeds periodic temporal signals into Gaussian representations for unified dynamic-static modeling. DeformableGS~\cite{yang2024deformable} introduces learnable deformation fields by mapping scenes into a canonical space and predicting time-dependent deformations. S$^3$Gaussian~\cite{huang2024textit} further transfers self-supervised dynamic decomposition to the 3DGS framework through 4D consistency constraints. Recent methods show that explicit 3D supervision can further improve the quality of dynamic reconstruction by decomposing scenes into foreground instances and static backgrounds. DrivingGaussian~\cite{zhou2024drivinggaussian} introduces incremental static Gaussians and composite dynamic Gaussian graphs for surround-view driving scenes. Street Gaussians~\cite{yan2024street} models time-varying appearances of dynamic vehicles with 4D spherical harmonics and jointly optimizes tracked poses. AutoSplat~\cite{khan2025autosplat} improves foreground reconstruction under sparse views using geometric symmetry priors and LiDAR initialization. HUGS~\cite{zhou2024hugs} incorporates semantic labels into Gaussian representations for joint reconstruction and scene understanding. OmniRe~\cite{chen2024omnire} further extends modeling targets to vehicles, pedestrians, and cyclists, combining rigid transforms, SMPL-based deformations, and self-supervised deformation fields within a unified scene graph framework. Despite their strong reconstruction quality, these methods still rely on per-scene optimization, which requires substantial training time for each scene and lacks cross-scene generalization.

\paragraph{Feedforward Reconstruction.} To overcome the efficiency bottleneck of per-scene optimization, feedforward reconstruction methods are trained on large-scale datasets to enable cross-scene generalization and directly predict 3D Gaussian representations in a single forward pass, reducing reconstruction time from hours to seconds. In general scene reconstruction, pixelSplat~\cite{charatan2024pixelsplat} and MVSplat~\cite{chen2024mvsplat} establish the basic paradigm of feedforward 3DGS through two-view encoding and multi-view cost volumes, respectively. DepthSplat~\cite{xu2025depthsplat} further improves geometric accuracy by jointly modeling depth estimation and Gaussian prediction. VGGT~\cite{wang2025vggt} proposes alternating spatial-temporal attention within a unified Transformer backbone to predict depth, point clouds, and camera parameters jointly. For driving scenes, DrivingForward~\cite{tian2025drivingforward} predicts Gaussians independently for each image and aggregates them into 3D space, serving as an early feedforward method for surround-view driving scenarios. STORM~\cite{yang2024storm} extends feedforward reconstruction to large-scale dynamic outdoor scenes by jointly predicting pixel-aligned Gaussians and motion fields in a self-supervised manner. DGGT~\cite{chen2025dggt} further removes the reliance on calibrated cameras by jointly predicting Gaussian representations and camera parameters, while introducing lifespan modeling for dynamic visibility over arbitrary-length sequences. Most existing feedforward methods still adopt pixel-aligned Gaussian prediction, in which Gaussians are decoded independently from image pixels, often leading to redundant primitives and limited cross-view consistency. To alleviate this issue, recent works have begun to explore pixel-unaligned paradigms. SparseSplat~\cite{zhang2026sparsesplat} reduces Gaussian counts through entropy-based adaptive sampling, while C3G~\cite{an2025c3g} introduces learnable query tokens with multi-view cross-attention to predict compact Gaussian sets. However, these methods are mainly validated on small indoor scenes and have not yet demonstrated strong performance in large-scale driving environments.

\paragraph{Multi-View 3D Object Detection.} Multi-view 3D object detection provides an important source of inspiration for sparse query-based scene representation. Early methods are mainly based on dense bird's-eye-view (BEV) representations. BEVDet~\cite{huang2021bevdet} adopts lift-splat projection for view transformation, while BEVFormer~\cite{li2024bevformer} introduces deformable attention for BEV feature generation and spatial-temporal fusion. BEVDepth~\cite{li2023bevdepth} further improves detection accuracy with explicit depth supervision, and BEVStereo~\cite{li2023bevstereo} enhances geometric reasoning by incorporating temporal stereo cues. Another line of work performs detection with sparse object queries. PETR~\cite{liu2022petr} utilizes 3D positional encoding and global cross-attention for query-based detection, while DETR3D~\cite{wang2022detr3d} directly samples multi-view image features using sparse 3D reference points. Building upon this paradigm, Sparse4D~\cite{lin2022sparse4d} introduces sparse spatial-temporal fusion, in which a compact set of 3D queries interacts across views and time to achieve efficient and accurate dynamic object detection. Inspired by these advances, our method adopts sparse 3D queries from object detection to feedforward 3D scene reconstruction.

\begin{figure}
  \centering
  \includegraphics[width=1.0\textwidth]{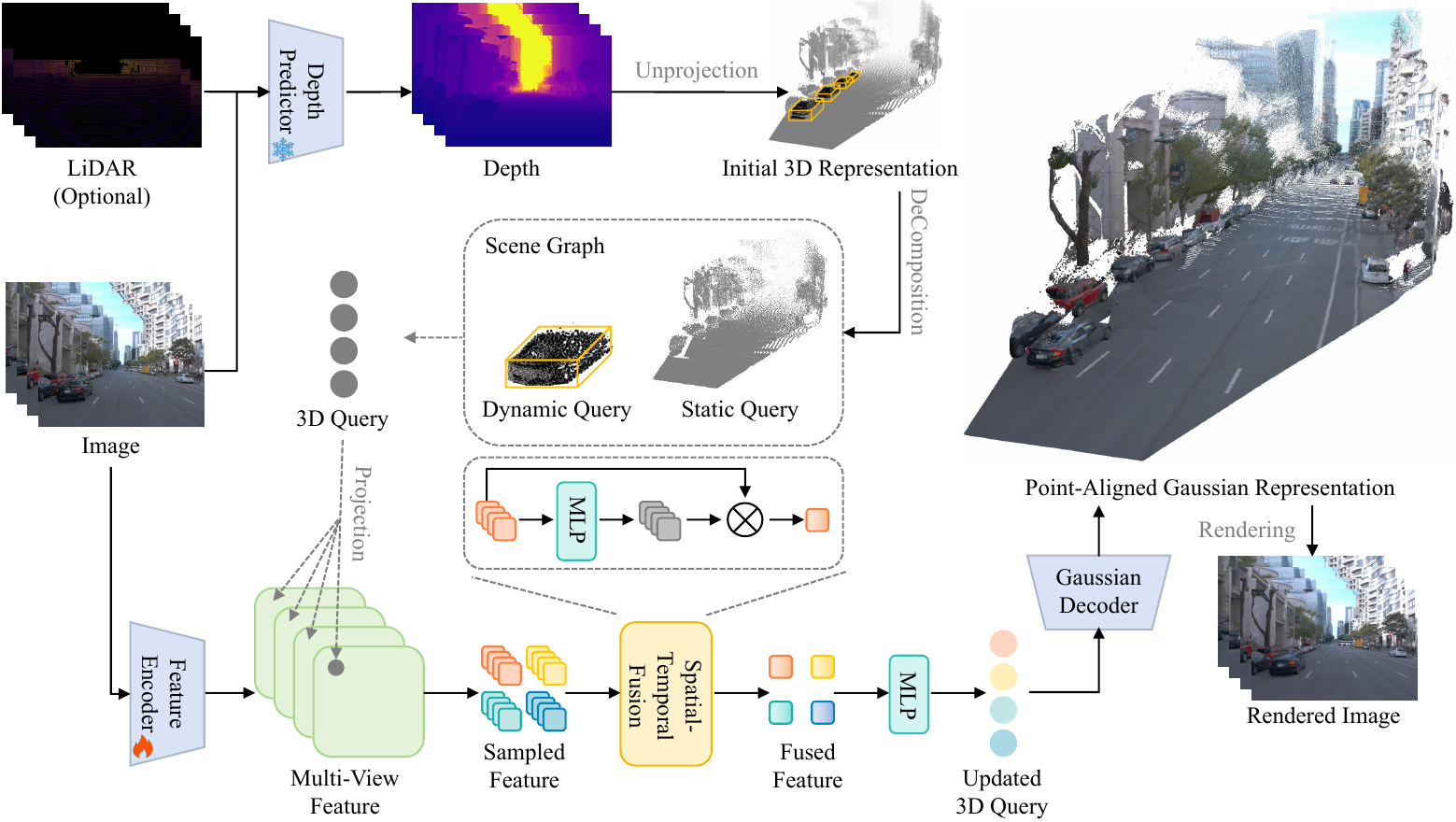}
  \caption{\textbf{Overview of PointForward.} We initialize sparse 3D queries in world space and project them onto multi-view image planes to aggregate features and geometric cues. A spatial-temporal fusion module produces coherent point-aligned representations, while scene-graph-based dynamic modeling ensures instance-level motion consistency for rendering dynamic driving scenes.}
  \label{fig:overview}
\end{figure}

\section{Method}

We first introduce how PointForward generates a point-aligned representation (Section ~\ref{subsec:query}). To handle dynamic scenes, we introduce a scene graph that directly models dynamics in 3D space (Section ~\ref{subsec:dynamic}). Dynamic queries are merged into canonical spaces, ensuring that each moving instance is globally unique across views. To achieve multi-view consistency, spatial-temporal fusion aggregates multi-view information onto each query, yielding a globally coherent 3D representation (Section ~\ref{subsec:fusion}). An overview of the model architecture is shown in Figure ~\ref{fig:overview}.

\subsection{Point-Aligned Representation}
\label{subsec:query}

Given a set of input images $\mathcal{I} = \{I_i \mid i=1,\dots,V \}$ with each $I_i \in \mathbb{R}^{H \times W \times 3}$. Each image has optional sparse LiDAR depth $D_i^L \in \mathbb{R}^{H \times W}$, camera extrinsics $[\mathbf{R}_i \mid \mathbf{t}_i]$, and camera intrinsics $\mathbf{K}_i$. We employ a pre-trained feature extractor (EfficientNetV2~\cite{tan2021efficientnetv2}) and a pre-trained depth estimation model (InfiniDepth~\cite{yu2026infinidepth}) to obtain image features $\mathbf{F}_i \in \mathbb{R}^{H \times W \times C}$ and dense depth $\mathbf{D}_i \in \mathbb{R}^{H \times W}$, respectively. During training, the feature extractor is fine-tuned to adapt to the reconstruction task, while the depth estimation model remains frozen to preserve its monocular prior. 

We then generate an initial dense query set $\mathcal{Q}_0 = \{ \mathbf{q}_{i,j} \mid i=1,\dots,V,\; j=1,\dots,H \times W \}$. Each query $\mathbf{q}$ is a 12-dimensional vector comprising a 3D spatial coordinate, a 3-channel RGB color, and a 6-dimensional Plücker coordinate~\cite{plucker1865xvii}. The 3D coordinate and Plücker coordinate of each query can be computed directly from the corresponding pixel depth and camera parameters. In essence, $\mathcal{Q}_0$ lifts image pixels into 3D space. At this stage, queries lack dynamic modeling and spatial-temporal fusion and therefore do not yet exhibit multi-view consistency.

Next, we perform dynamic modeling using scene graphs, with detailed procedures referred to Section ~\ref{subsec:dynamic}. After this step, the initial query set $\mathcal{Q}_0$ is decomposed into a static query set $\mathcal{Q}_s = \{ \mathbf{q}_{i,j} \mid i=1,\dots,V,\; j=1,\dots,N_s \}$ and a dynamic query set $\mathcal{Q}_d = \{ \mathbf{q}_{i,j} \mid i=1,\dots,O,\; j=1,\dots,N_d \}$, where $O$ denotes the number of dynamic instances in the scene. To reduce computation and avoid redundancy, we sample $N_s$ static queries per view and $N_d$ dynamic queries per instance. This yields our final point-aligned representation $\mathcal{Q} = \mathcal{Q}_s \cup \mathcal{Q}_d = \{\mathbf{q}_i\}_{i=1}^{N}$, where $N = V * N_s + O * N_d$ denotes the total number of sparse queries. It subsequently undergoes spatial-temporal fusion as detailed in Section ~\ref{subsec:fusion}.

\subsection{Dynamic Modeling with Scene Graph}
\label{subsec:dynamic}

To model scene dynamics, we first obtain 3D bounding boxes for all moving instances. Let $O$ be the number of dynamic instances. For each instance $i$, there is a set of bounding boxes $\mathcal{B}_i = \{ \mathbf{b}_{i,j} \}_{j=1}^{B_i}$, where $B_i$ is the number of available boxes. These boxes are temporally ordered, with $\mathbf{b}_{i,1}$ corresponding to the earliest timestamp. We designate $\mathbf{b}_{i,1}$ as the canonical reference box for instance $i$, whose local coordinate frame establishes the canonical space. Each box $\mathbf{b}_{i,j}$ is parameterized in world space by a center $\mathbf{c}_{i,j} \in \mathbb{R}^3$, dimensions $\mathbf{m}_{i,j} \in \mathbb{R}^3$, and a rotation matrix $\mathbf{R}_{i,j}^{\mathcal{B}}$. A 3D point $\mathbf{p} \in \mathbb{R}^3$ lies inside $\mathbf{b}_{i,j}$ if the following condition holds:
\begin{equation}
\bigl| (\mathbf{p} - \mathbf{c}_{i,j}) (\mathbf{R}_{i,j}^{\mathcal{B}})^\top \bigr| \preceq \frac{\mathbf{m}_{i,j}}{2},
\end{equation}
where $\preceq$ denotes element-wise comparison.

Specifically, for a query $\mathbf{q}\in\mathcal{Q}_0$ with 3D spatial coordinate $\mathbf{p}$, if $\mathbf{p}$ lies inside any bounding box $\mathbf{b}_{i,j}$, $\mathbf{q}$ is assigned to instance $i$ and marked as a dynamic query. Otherwise, it is treated as a static query belonging to the background. For each dynamic query associated with instance $i$, we further transform its world-space coordinate into the canonical space defined by the reference box $\mathbf{b}_{i,1}$:
\begin{equation}
\tilde{\mathbf{p}} = (\mathbf{p} - \mathbf{c}_{i,j}) (\mathbf{R}_{i,j}^{\mathcal{B}})^\top \mathbf{R}_{i,1}^{\mathcal{B}} + \mathbf{c}_{i,1},
\end{equation}
where $\tilde{\mathbf{p}}$ denotes the canonicalized coordinate. After canonicalization, all dynamic queries belonging to the same instance are represented in a consistent, instance-centric coordinate frame, while static queries remain in the global world space. This scene graph formulation disentangles dynamic instance motion from static background geometry and provides structured motion priors for the subsequent spatial-temporal fusion.

Consequently, the initial dense query set $\mathcal{Q}_0$ is decomposed into a static query set $\mathcal{Q}_s$ and a dynamic query set $\mathcal{Q}_d$, as described in Section ~\ref{subsec:query}, where sparse sampling is applied to reduce redundancy and computational cost.

\subsection{Spatial-Temporal Fusion}
\label{subsec:fusion}

We first obtain a set of dynamic masks $\mathcal{M}=\{M_i \mid i=1,\dots, V\}$ by projecting all bounding boxes onto each image plane, where each $M_i \in \mathbb{R}^{H \times W}$ indicates dynamic regions in view $i$.  For each $\mathbf{q}\in\mathcal{Q}$, we project its 3D spatial coordinate onto every input image $I_i$ and sample the corresponding feature $\mathbf{f}_i \in \mathbb{R}^C$ from $\mathbf{F}_i$. Dynamic queries are first transformed from the canonical space back to world space before being projected. To avoid inconsistent motion features, if a static query falls inside a dynamic region in $M_i$, the sampled feature is set to zero. Besides $\mathbf{f}_i$, we also compute the absolute depth difference $d_i$ between the query depth and the corresponding pixel depth from $\mathbf{D}_i$. This value reflects the geometric consistency of $\mathbf{f}_i$, where a smaller $d_i$ indicates a more reliable observation. We then fuse the sampled feature $\mathbf{f}_i$ and depth difference $d_i$ as:
\begin{equation}
\mathbf{f}_{s,i} = \phi \left( \left[ \phi(\mathbf{f}_i) ; \phi(d_i) \right] \right),
\end{equation}
where $\phi(\cdot)$ denotes a multi-layer MLP operation applying to the feature dimension.

We further incorporate temporal cues for dynamic scenes. Let $t_i$ denote the normalized timestamp of image $I_i$, and let $t$ be the timestamp associated with query $\mathbf{q}$. The initial associated feature of $\mathbf{q}$ is denoted as $\mathbf{f}_q$, and compute the fusion weight for the $i$-th view as:
\begin{equation}
\begin{aligned}
    \mathbf{f}_{t,i} &= \left[\mathbf{f}_q \odot \mathbf{f}_{s,i};t == t_i;t\right], \\
    \omega_i &= \operatorname{Softmax}\bigl(\phi(\mathbf{f}_{t,i})\bigr),
\end{aligned}
\end{equation}
where $\odot$ denotes element-wise multiplication. We then obtain the fused feature $\mathbf{f}$ and lifespan parameter $\sigma$ as:
\begin{equation}
\begin{aligned}
    \sigma &= \phi_v \left( \left[ \phi_c(\mathbf{f}_{t,1}), \dots, \phi_c(\mathbf{f}_{t,V}) \right] \right), \\
    \mathbf{f} &= \sum_{i=1}^{V} \omega_i \mathbf{f}_{t,i},
\end{aligned}
\end{equation}
where $\phi_c(\cdot)$ denotes a multi-layer MLP that reduces the feature dimension of $\mathbf{f}_{t,i}$ to a scalar, and $\phi_v(\cdot)$ denotes another multi-layer MLP that aggregates across views to produce the final lifespan parameter $\sigma$. Finally, we obtain the updated query representation by fusing all available information as:
\begin{equation}
\mathbf{q}' = \phi \left( \left[ \phi(\mathbf{q}); \phi(\mathbf{f}); \phi(t); \phi(\sigma) \right] \right).
\end{equation}

\subsection{Rendering}

We decode each $\mathbf{q}'$ into a 3D Gaussian representation $\mathbf{g} \equiv (\mathbf{c}, \boldsymbol{\mu}, \mathbf{r}, \mathbf{s}, o)$, where the color feature $\mathbf{c} \in \mathbb{R}^{N_c}$, center $\boldsymbol{\mu} \in \mathbb{R}^{3}$, rotation quaternion $\mathbf{r} \in \mathbb{R}^{4}$, scale $\mathbf{s} \in \mathbb{R}^{3}$, and opacity $o \in \mathbb{R}$ are predicted from $\mathbf{q}'$. The opacity of each Gaussian at another timestamp $t'$ is adjusted as:
\begin{equation}
o_{t'} = o \cdot e^{-\frac{1}{2}\bigl(\frac{t' - t}{\sigma + 1}\bigr)^2}.
\end{equation}

When rendering at timestamp $t'$, Gaussians corresponding to dynamic queries are transformed from the canonical space back to the world space of the target timestamp. Following recent works such as UniSim~\cite{yang2023unisim} and SplatAD~\cite{hess2025splatad}, we represent each Gaussian by a high-dimensional color feature $\mathbf{c}$ rather than directly predicting RGB values. As a result, Gaussian rendering produces a feature map rather than a color image. We then employ a lightweight multi-layer UNet to map the rendered feature map into three channels, yielding the final rendered RGB image.

\subsection{Loss Function}

We supervise the rendered images using a combination of pixel-wise $\ell_1$ and perceptual losses. Specifically, the overall training loss function is defined as:
\begin{equation}
\mathcal{L} = \mathcal{L}_{\ell_1} + \lambda \mathcal{L}_{\text{LPIPS}},
\end{equation}
where $\mathcal{L}_{\ell_1}$ denotes the $\ell_1$ reconstruction loss and $\mathcal{L}_{\text{LPIPS}}$ denotes the perceptual LPIPS loss.

\section{Experiment}

\subsection{Implementation Detail}
We implement our PointForward using PyTorch and train it on 8 NVIDIA H20 GPUs. Training runs for 45K iterations with a global batch size of 8 and a learning rate of $5 \times 10^{-4}$. The training process uses the AdamW optimizer with a cosine learning rate scheduler, which includes a linear warm-up phase over the first 2K iterations. We use a image size of $576 \times 768$ and multi-resolution rendering supervision, randomly sampled from $350 \times 518$, $576 \times 768$, and $700 \times 1036$. We train PointForward on Waymo from scratch and fine-tune it on nuScenes from Waymo's checkpoint. Parameters are set as $N_s$ = 100K, $N_d$ = 10K, $N_c$ = 128, $\lambda$ = 0.2.

\subsection{Dataset and Metric}
We primarily conduct experiments on the Waymo Open Dataset~\cite{sun2020scalability}, which contains 798 training and 202 validation sequences. Each sequence consists of a 20-second video captured at 10 FPS. During training, we randomly sample 20 frames (2 s) from the frontal three cameras for supervision, and the context frames are selected at 0 s, 0.5 s, 1.0 s, and 1.5 s within the sampled clip. During validation, we follow the same setup with DGGT~\cite{chen2025dggt}. Context frames are inputs, and the model predicts the remaining frames to validate rendering performance. We further evaluate zero-shot generalization and fine-tuning performance on the nuScenes dataset~\cite{caesar2020nuscenes}, which contains 750 training and 150 validation sequences. The clip sampling strategy follows the same protocol as Waymo. Longer video clip results are provided in the supplementary video.

We mainly report PSNR, SSIM, and Depth RMSE (D-RMSE) as the primary evaluation metrics. For extrapolated-view experiments, we also report FID to measure the distributional distance between rendered and ground-truth images.

\subsection{Comparison with SOTA Method}

\paragraph{Baseline.} We compare our method with a range of existing approaches, including state-of-the-art feedforward pixel-aligned methods such as STORM~\cite{yang2024storm} and DGGT~\cite{chen2025dggt}. For a fair comparison, we disable the diffusion-based rendering refinement module in DGGT for all extrapolated-view experiments and qualitative visualizations. This module is not specific to DGGT and can be applied to any method to refine rendered images. Removing it allows for a direct comparison of the raw Gaussian rendering quality across methods.

\begin{table}[t]
    \centering
    \caption{\textbf{Quantitative results on Waymo.} We compare image quality and geometry metrics against both per-scene optimization and feedforward methods. PSNR, SSIM, and D-RMSE are reported.}
    \label{tab:waymo}
    \begin{tabular}{lcccccc}
        \hline
        \multirow{2}{*}{Method} & \multicolumn{3}{c}{Dynamic Only} & \multicolumn{3}{c}{Full Image} \\
         & PSNR↑ & SSIM↑ & D-RMSE↓ & PSNR↑ & SSIM↑ & D-RMSE↓ \\ \hline
        EmerNeRF~\cite{yang2023emernerf} & 17.79 & 0.255 & 40.88 & 24.51 & 0.738 & 33.99 \\
        3DGS~\cite{kerbl20233d} & 17.13 & 0.267 & 13.88 & 25.13 & 0.741 & 19.68 \\
        PVG~\cite{chen2026periodic} & 15.51 & 0.128 & 15.91 & 22.38 & 0.661 & 13.01 \\
        DeformableGS~\cite{yang2024deformable} & 17.10 & 0.266 & 12.14 & 25.29 & 0.761 & 14.79 \\
        LGM~\cite{tang2024lgm} & 19.58 & 0.443 & 9.43 & 23.59 & 0.691 & 8.02 \\
        GS-LRM~\cite{zhang2024gs} & 20.02 & 0.520 & 9.95 & 25.18 & 0.753 & 7.94 \\
        Depth Anything 3~\cite{lin2025depth} & 15.96 & 0.429 & 12.20 & 23.27 & 0.686 & 11.66 \\
        AnySplat~\cite{jiang2025anysplat} & 15.99 & 0.418 & 10.45 & 23.32 & 0.687 & 10.31 \\
        STORM~\cite{yang2024storm} & 22.10 & 0.624 & 7.50 & 26.38 & 0.794 & 5.48 \\
        DGGT~\cite{chen2025dggt} & 22.80 & 0.651 & 6.55 & 27.41 & 0.846 & 3.47 \\
        Ours & \textbf{25.01} & \textbf{0.768} & \textbf{6.07} & \textbf{28.48} & \textbf{0.861} & \textbf{3.35} \\ \hline
    \end{tabular}
\end{table}

\begin{figure}[ht]
  \centering
  \includegraphics[width=1.0\textwidth]{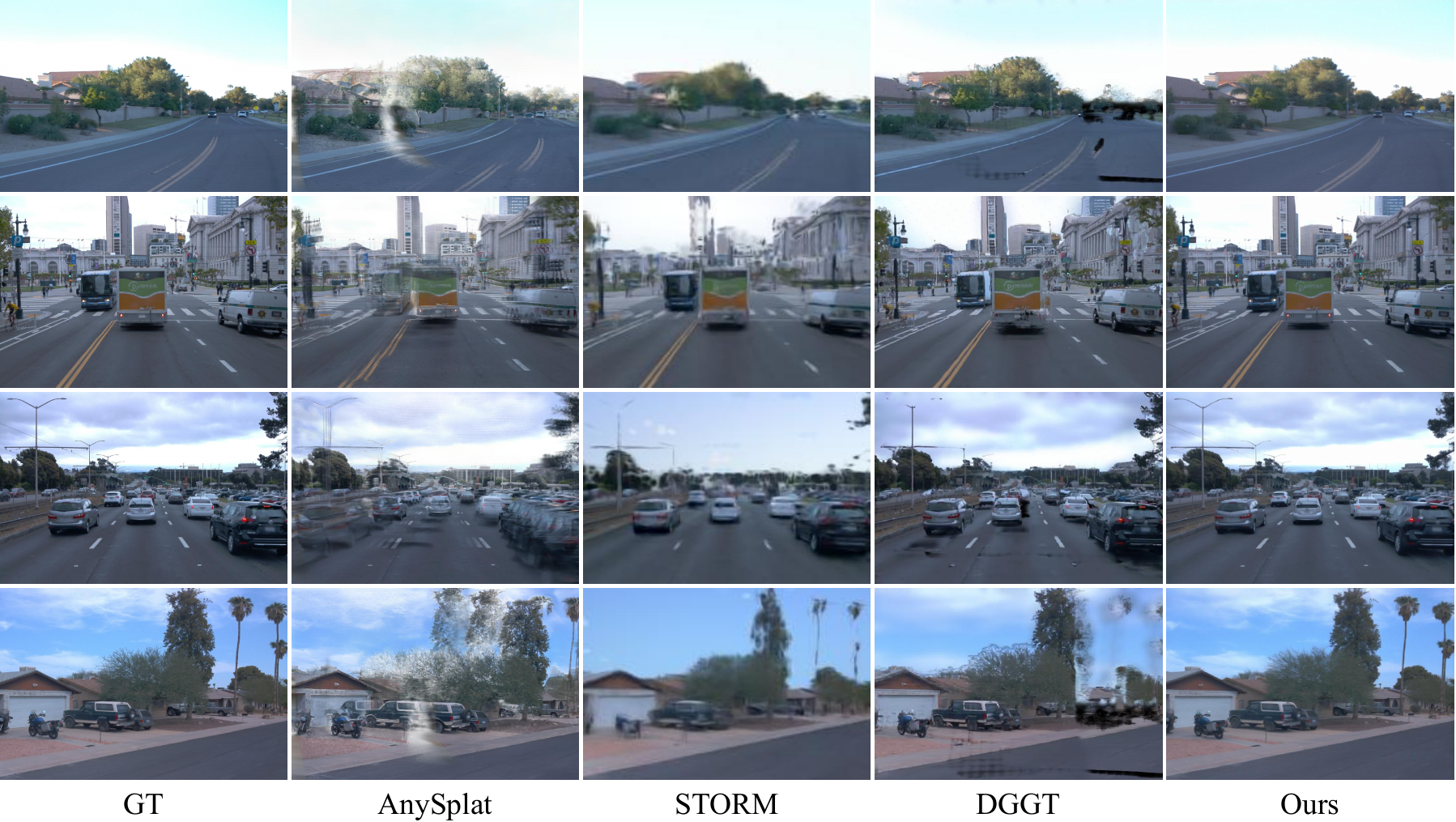}
  \caption{\textbf{Qualitative results on Waymo.} PointForward produces sharper structures and cleaner object boundaries with significantly reduced ghosting artifacts.}
  \label{fig:waymo}
\end{figure}

\paragraph{Waymo.} As shown in Table ~\ref{tab:waymo}, PointForward achieves state-of-the-art performance on both the full image and dynamic regions. In particular, our method surpasses DGGT by 1.1 dB in PSNR on the full image and by 2.2 dB on dynamic regions, demonstrating the advantage of our point-aligned representation in enforcing multi-view consistency and improving dynamic reconstruction quality. As illustrated in Figure ~\ref{fig:waymo}, our method produces sharper structures and cleaner object boundaries with significantly reduced ghosting artifacts. In contrast, AnySplat struggles to handle dynamic instances and large viewpoint changes, resulting in severe artifacts. STORM is limited by its training resolution and tends to generate over-smoothed results with missing fine details. DGGT often introduces ghosting in high-frequency regions and exhibits unstable rendering during rapid viewpoint changes due to its less coherent modeling.

We further evaluate reconstruction using only context frames. As shown in Table ~\ref{tab:reconstruction}, PointForward outperforms DGGT by 3 dB in PSNR and nearly 0.2 in SSIM. Even when restricted to render context frames, PointForward still consistently outperforms prior methods, highlighting the effectiveness of our point-aligned Gaussian representation. This suggests that improved performance stems from a more coherent and expressive Gaussian representation.

\begin{table}[t]
    \begin{minipage}[t]{0.39\textwidth}
        \centering
        \caption{\textbf{Quantitative reconstruction results on Waymo.}}
        \label{tab:reconstruction}
            \begin{tabular}{lccc}
                \hline
                Method & PSNR↑ & SSIM↑ \\ \hline
                AnySplat~\cite{jiang2025anysplat} & 24.59 & 0.730 \\
                STORM~\cite{yang2024storm} & 26.55 & 0.851 \\
                DGGT~\cite{chen2025dggt} & 30.54 & 0.884  \\
                Ours & \textbf{33.96} & \textbf{0.951} \\ \hline
            \end{tabular}
    \end{minipage}
    \hfill
    \begin{minipage}[t]{0.59\textwidth}
        \centering
        \caption{\textbf{Quantitative results on nuScenes.} We achieve SOTA performance on zero-shot and fine-tuning.}
        \label{tab:nuscenes}
            \begin{tabular}{lcccc}
                \hline
                \multirow{2}{*}{Method} & \multicolumn{2}{c}{Zero-Shot} & \multicolumn{2}{c}{Fine-Tuning} \\
                 & PSNR↑ & SSIM↑ & PSNR↑ & SSIM↑ \\ \hline
                STORM~\cite{yang2024storm} & 17.77 & 0.669 & 24.54 & 0.784 \\
                DGGT~\cite{chen2025dggt} & 25.31 & 0.794 & 26.63 & 0.813 \\
                Ours & \textbf{26.54} & \textbf{0.821} & \textbf{27.50} & \textbf{0.826} \\ \hline
            \end{tabular}
    \end{minipage}
\end{table}

\begin{figure}[ht]
  \centering
  \includegraphics[width=1.0\textwidth]{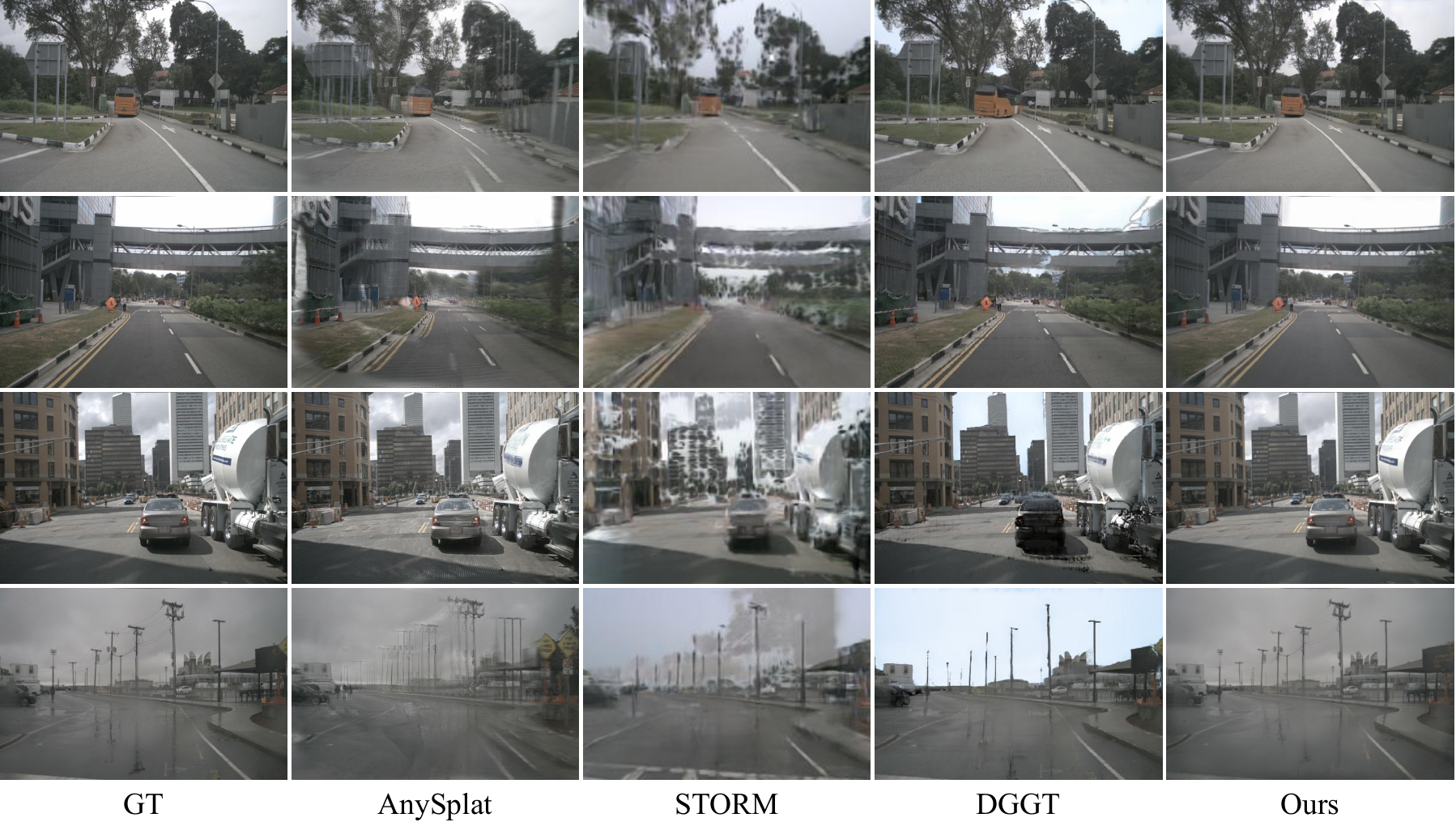}
  \caption{\textbf{Qualitative zero-shot results on nuScenes.}}
  \label{fig:nuscenes}
\end{figure}

\paragraph{nuScenes.} We evaluate PointForward on nuScenes under both zero-shot and fine-tuning settings to assess its generalization and adaptation capabilities. As shown in Table ~\ref{tab:nuscenes}, PointForward consistently outperforms prior methods in both settings, with more significant gains in the zero-shot scenario, achieving +1.2 dB in PSNR and +0.3 in SSIM. This demonstrates that our point-aligned representation is more stable and generalizable than pixel-aligned counterparts. As illustrated in Figure ~\ref{fig:nuscenes}, prior methods often produce incorrect geometry and motion when applied to unseen scenarios. In contrast, PointForward accurately reconstructs scene structures and captures instance motions, even under cross-dataset conditions.

\begin{figure}[ht]
  \centering
  \includegraphics[width=1.0\textwidth]{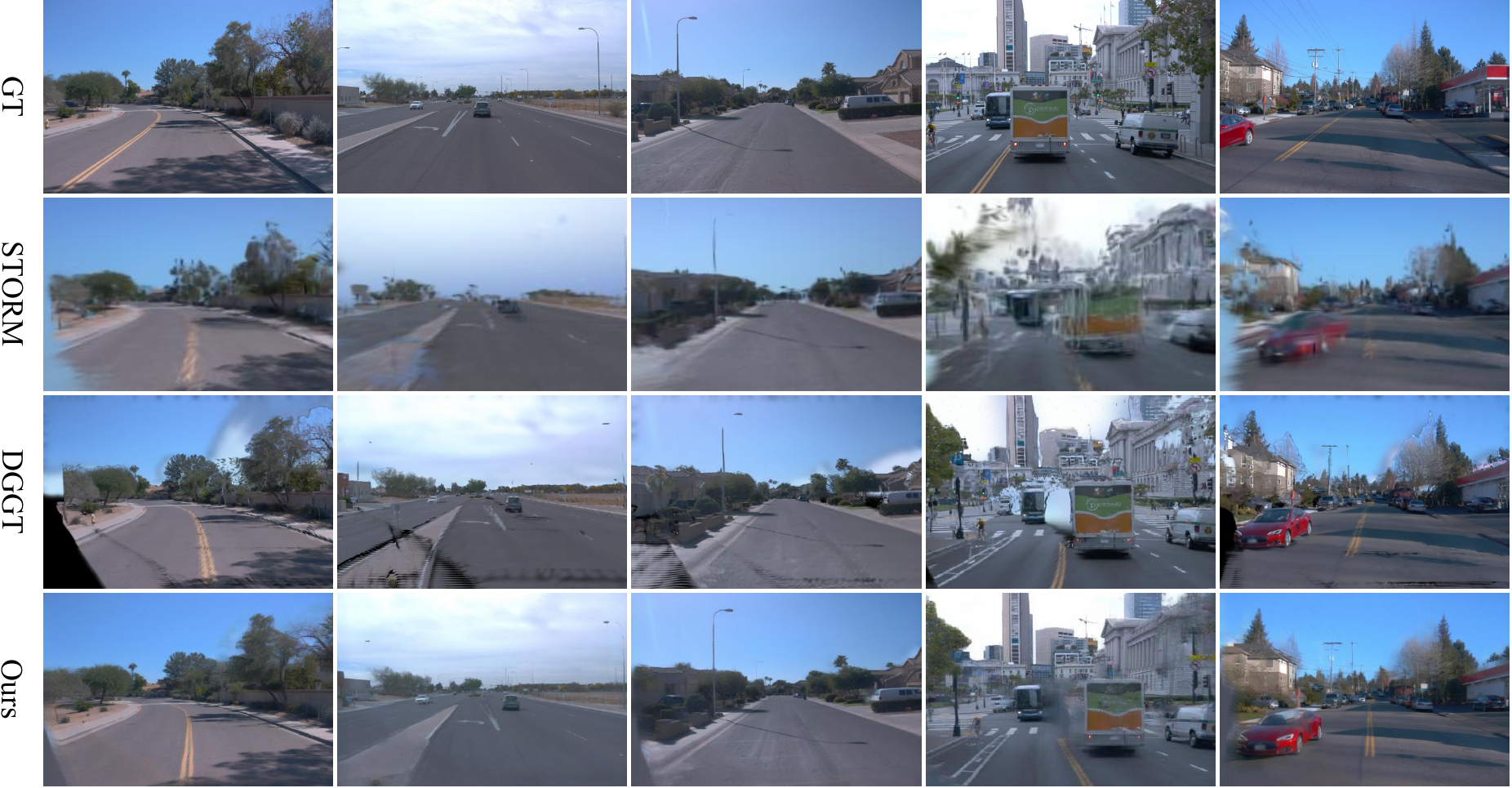}
  \caption{\textbf{Qualitative extrapolated-view results on Waymo.} The camera is left shifted 3 meters.}
  \label{fig:shift}
\end{figure}

\paragraph{Extrapolated View.} Since previous experiments were conducted on interpolated views, we further conduct extrapolated-view experiments to validate the effectiveness of point-aligned representation. For fair comparison, all methods take a single front-view camera as input. Note that DGGT predicts relative camera parameters and relative depth; therefore, we align its predicted depth with ground-truth depth to approximate absolute camera motion, which may introduce minor inaccuracies. As shown in Table ~\ref{tab:shift}, PointForward significantly outperforms STORM and DGGT in FID. Qualitative results in Figure ~\ref{fig:shift} further demonstrate that our method preserves correct geometric structures in extrapolated viewpoints, such as lane layouts and vehicle shapes. These results highlight that point-aligned representation maintains stronger geometric consistency than pixel-aligned representation.

\subsection{Ablation Study}

\paragraph{Effectiveness of spatial-temporal fusion.} As shown in Table ~\ref{tab:ablation}, directly averaging the sampled multi-view information instead of applying spatial-temporal fusion leads to a noticeable performance drop. Removing depth difference information also degrades image quality. These results demonstrate the necessity of our spatial-temporal fusion module and the importance of incorporating geometric consistency cues during feature aggregation.

\begin{table}[t]
    \begin{minipage}[t]{0.49\textwidth}
        \centering
        \caption{\textbf{Quantitative extrapolated-view results on Waymo.} Lower FID is better.}
        \label{tab:shift}
        \begin{tabular}{cccc}
            \hline
            Shift & STORM~\cite{yang2024storm} & DGGT~\cite{chen2025dggt} & Ours \\ \hline
            1 m & 47.31  & 49.51 & \textbf{16.95} \\
            2 m & 63.09 & 61.35 & \textbf{25.80} \\
            3 m & 81.54 & 77.51 & \textbf{33.71} \\ 
            4 m & 102.19 & 96.96 & \textbf{42.63} \\ 
            6 m & 140.17 & 139.79 & \textbf{65.99} \\ \hline
        \end{tabular}
    \end{minipage}
    \hfill
    \begin{minipage}[t]{0.49\textwidth}
        \centering
        \caption{\textbf{Ablation Study on Waymo.} w/o LiDAR depth still achieves SOTA performance.}
        \label{tab:ablation}
        \begin{tabular}{lcc}
            \hline
            Method & PSNR↑ & SSIM↑ \\ \hline
            w/o weighted fusion & 28.01 & 0.851 \\
            w/o depth-aware fusion & 28.28 & 0.855 \\
            w/o lifespan & 27.56  & 0.840 \\
            w/o LiDAR depth & 28.00 & 0.847 \\
            Ours & \textbf{28.48} & \textbf{0.861} \\ \hline
        \end{tabular}
    \end{minipage}
\end{table}

\paragraph{Influence of lifespan.} As shown in Table ~\ref{tab:ablation}, removing the lifespan parameter results in only a 0.92 dB PSNR drop. In contrast, DGGT reports a much larger degradation of 3.11 dB. This suggests that our point-aligned representation is inherently more global and coherent, while the pixel-aligned representation relies more heavily on additional temporal parameters to model scene dynamics.

\paragraph{Independence of LiDAR depth.} As shown in Table ~\ref{tab:ablation}, PointForward still achieves 28.00 PSNR and 0.847 SSIM without using LiDAR depth, by directly adopting depths predicted from an external monocular depth estimation model~\cite{xu2026pixel2, xu2026pixel} such as MoGe-2~\cite{wang2025moge}. The performance still surpasses STORM and DGGT, achieving state-of-the-art results. This demonstrates that our method does not rely on LiDAR and can generalize effectively with only image-based geometric priors.

\paragraph{Query sampling and budget sensitivity.} We uniformly sample static queries within each view and dynamic queries within each instance, without introducing texture-, or category-specific sampling heuristics. To study the sensitivity to the query budget, we independently train PointForward with different numbers of sampled static and dynamic queries while keeping all other settings unchanged. As shown in Table~\ref{tab:query}, halving both $N_s$ and $N_d$ reduces PSNR by only 0.03 dB and SSIM by 0.001. Even when trained with only one quarter of the default query budget, PointForward still achieves 28.25 PSNR and 0.856 SSIM. These results show that PointForward is robust to the choice of query budget and can maintain competitive reconstruction quality with substantially fewer queries.

\paragraph{Effect of image encoder and initialization.} We further study the influence of the image encoder architecture and its initialization. As shown in Table~\ref{tab:encoder}, replacing the pretrained EfficientNetV2-S~\cite{tan2021efficientnetv2} with a randomly initialized counterpart results in only a small decrease from 28.48 to 28.40 PSNR. Using a pretrained ResNet-50~\cite{he2016deep} also achieves comparable performance of 28.42 PSNR and 0.859 SSIM. These results indicate that PointForward is robust to the choice and initialization of the image encoder, while pretrained EfficientNetV2-S provides the best overall performance and is therefore adopted as our default setting.

\paragraph{Robustness to predicted 3D box trajectories.} Our main experiments use ground-truth 3D box trajectories to isolate the reconstruction capability of different methods. We further evaluate PointForward with predicted trajectories to assess its robustness under practical settings. On Waymo, we retain GT trajectories during training and replace all test-time 3D boxes with detector-and-tracker predictions~\cite{yan2024street}. As shown in Table~\ref{tab:track}, this results in only a 0.30 dB PSNR, 0.005 SSIM, and 0.51 dB dynamic PSNR decrease. PointForward with predicted trajectories still substantially outperforms STORM and DGGT, particularly on dynamic regions.

On nuScenes, we use predicted box trajectories~\cite{lin2023sparse4d} during both training and inference, completely removing human-annotated dynamic trajectories from the pipeline. Despite detection, tracking, and temporal interpolation errors, performance decreases by only 0.25 dB in PSNR, 0.003 in SSIM, and 0.45 dB in dynamic PSNR. These results show that our scene-graph-based dynamic modeling is robust to trajectory prediction errors and does not require human-annotated trajectories at inference.

\begin{table}[t]
    \begin{minipage}[t]{0.39\textwidth}
        \centering
        \caption{\textbf{Query budget sensitivity on Waymo.}
        PointForward is robust to different numbers of sampled queries.}
        \label{tab:query}
        \begin{tabular}{ccc}
            \hline
            $N_s / N_d$ & PSNR$\uparrow$ & SSIM$\uparrow$ \\ \hline
            25K / 2.5K & 28.25 & 0.856 \\
            50K / 5K   & 28.45 & 0.860 \\
            100K / 10K & \textbf{28.48} & \textbf{0.861} \\ \hline
        \end{tabular}
    \end{minipage}
    \hfill
    \begin{minipage}[t]{0.59\textwidth}
        \centering
        \caption{\textbf{Effect of image encoder and initialization on Waymo.}
        The pretrained EfficientNetV2-S encoder gives the best performance.}
        \label{tab:encoder}
        \begin{tabular}{l lcc}
            \hline
            Image encoder & Initialization & PSNR$\uparrow$ & SSIM$\uparrow$ \\ \hline
            EfficientNetV2-S & Pretrained & \textbf{28.48} & \textbf{0.861} \\
            EfficientNetV2-S & Random     & 28.40 & 0.855 \\
            ResNet-50        & Pretrained & 28.42 & 0.859 \\ \hline
        \end{tabular}
    \end{minipage}
\end{table}

\begin{table}[t]
    \centering
    \caption{\textbf{Robustness to predicted 3D box trajectories.}
    PointForward remains robust when replacing ground-truth box trajectories
    with detector-and-tracker predictions.}
    \label{tab:track}
    \begin{tabular}{llccc}
        \hline
        Dataset / Method & Box Trajectory
        & PSNR$\uparrow$ & SSIM$\uparrow$ & Dynamic PSNR$\uparrow$ \\ \hline

        Waymo -- STORM~\cite{yang2024storm}
        & None
        & 26.38 & 0.794 & 22.10 \\

        Waymo -- DGGT~\cite{chen2025dggt}
        & None
        & 27.41 & 0.846 & 22.80 \\

        Waymo -- PointForward
        & GT
        & \textbf{28.48} & \textbf{0.861} & \textbf{25.01} \\

        Waymo -- PointForward
        & Predicted
        & 28.18 & 0.856 & 24.50 \\ \hline

        nuScenes -- PointForward
        & GT
        & \textbf{27.50} & \textbf{0.826} & \textbf{23.88} \\

        nuScenes -- PointForward
        & Predicted
        & 27.25 & 0.823 & 23.43 \\ \hline
    \end{tabular}
\end{table}

\section{Conclusion}

In this paper, we present PointForward, a feedforward driving reconstruction framework through point-aligned representations. PointForward introduces a point-aligned representation based on sparse 3D queries in world space, enabling explicit multi-view consistency and coherent global geometry. We further incorporate scene-graph-based dynamic modeling and spatial-temporal fusion to achieve consistent motion representation and robust feature aggregation across views and timestamps. Extensive experiments on Waymo and nuScenes demonstrate that PointForward achieves state-of-the-art performance and shows strong robustness under extrapolated viewpoints. We hope our work can inspire future research toward more scalable and consistent feedforward reconstruction frameworks.

\paragraph{Limitation.} Current dynamic modeling mainly focuses on rigid instances represented by 3D bounding boxes. As a result, PointForward is less effective for highly non-rigid motions, such as pedestrians. Extending point-aligned representations toward more flexible non-rigid dynamic modeling remains an important direction for future work.

\section{Acknowledgment}

We use publicly available datasets, models, and code resources, including Waymo, nuScenes, and other public research resources. We confirm that all such resources are used solely for academic research purposes and not for any commercial activity, in accordance with their respective licenses and terms of use. This research is supported by National Natural Science Foundation of China (U25B2045, 62472184), the Fundamental Research Funds for the Central Universities, and the Innovation Project of Optics Valley Laboratory (Grant No. OVL2025YZ005)

{
\bibliographystyle{plain}
\bibliography{reference}
}







\end{document}